\documentclass{article}

\usepackage{arxiv}

\usepackage[utf8]{inputenc} 
\usepackage[T1]{fontenc}    
\usepackage{hyperref}       
\usepackage{url}            
\usepackage{booktabs}       
\usepackage{amsfonts}       
\usepackage{nicefrac}       
\usepackage{microtype}      
\usepackage{lipsum}
\usepackage{amsmath}
\usepackage{cite}
\usepackage{graphicx}
\graphicspath{ {./images/} }

\title{LSTM recurrent neural network assisted aircraft stall prediction for enhanced situational awareness}

\author{
  Tahsin Sejat Saniat \\
  Mechanical and Production Engineering\\
  Islamic University of Technology\\
  Boardbazar, Gazipur 1704\\
  \texttt{tahsinsejat@iut-dhaka.edu} \\
   \And
 Tahiat Goni \\
  Electrical and Computer Engineering\\
  University of Alberta\\
  Edmonton, AB T6G 2R3 \\
  \texttt{tahiat@ualberta.ca}
  \And
  Shaikat M. Galib \\
  H2O.ai\\
  2307 Leghorn St, Mountain View, CA 94043 \\
  \texttt{shaikat.galib@h2o.ai}
}

\begin{document}
\maketitle
\begin{abstract}
Since the dawn of mankind’s introduction to powered flights, there have been multiple incidents which can be attributed to aircraft stalls. Most modern-day aircraft are equipped with advanced warning systems to warn the pilots about a potential stall, so that pilots may adopt the necessary recovery measures. But these warnings often have a short window before the aircraft actually enters a stall and require the pilots to act promptly to prevent it. In this paper, we propose a deep learning-based approach to predict an Impending stall, well in advance, even before the stall-warning is triggered. We leverage the capabilities of long short-term memory (LSTM) recurrent neural networks (RNN) and propose a novel approach to predict potential stalls from the sequential in-flight sensor data. Three different neural network architectures were explored. The neural network models, trained on 26400 seconds of simulator flight data are able to predict a potential stall with $>95\%$ accuracy, approximately 10 seconds in advance of the stall-warning trigger. This can significantly augment the Pilot's preparedness to handle an unexpected stall and will add an additional layer of safety to the traditional stall warning systems.
\end{abstract}


\section{Introduction}
Stall related incidents are still one of the most common challenges in the aviation industry. It generally occurs when the wings generate inadequate lift and can potentially result in Loss of Control-Inflight (LOC-I), which can be fatal if the pilot is unable to recover the aircraft in time. According to Boeing’s statistical summary of aviation accidents, 1183 out of a total 2532 fatalities from 2009 through 2018 were related to loss of control-inflight (LOC-I) ~\cite{airplanes2016statistical}. Stall is one of the main sources of LOC-I and accounted for nearly 20\% of all the annual fatal accidents between 2000-2014~\cite{collins2015stall}. Despite the intensive pilot training to recognize and handle stalls, it is prevalent as a major cause or co-factor in aviation accidents~\cite{aircraft201423rd} and demands efficient and innovative approach for detection and mitigation. Stall can occur due to a complex sequence of events leading to aircraft upset or due to adverse weather, clear air turbulence, or windshear in both visual (VMC) and instrument meteorological conditions (IMC)~\cite{BibEntry2020Nov}. The outcome of such an event is primarily dependent on the prompt response from the flight crew in taking recovery measures. According to the federal aviation administration (FAA), the likelihood of a full stall increases proportionally with the amount of time taken by the pilot to identify an impending stall and act upon it~\cite{f2016a}. In addition, the current stall identification strategy is reliant on the crew’s ability to infer a potential stall from the sight, sound, feel or the change in aircraft handling characteristics~\cite{f2016a}. However, this is not a foolproof method to prevent stall and is susceptible to failure as established by its frequent occurrences to date. During the training phase, recognition and recovery from stall is administered in a controlled and predictable environment, with the pilots being aware of a potential stall. However, in real life, pilots are often unaware as the aircraft is maneuvered into a stall unexpectedly, and have little time to address the sudden change in the cockpit environment. Therefore, technology that can assist pilots to predict potential stalls in-flight can inhibit this problem and improve situational awareness, thereby reducing flight envelope breach and stall related fatalities. 
LSTM recurrent neural networks, initially introduced by Hochreiter and Schmidhuber~\cite{hochreiter1997long} are very good in dealing with sequential data~\cite{dabek2015neural,lipton2015learning,zhao2017learning,jiao2018short} and since the changes in various flight parameters leading up to a stall are sequence data, this category of deep neural networks can perform exceptionally well in identifying potential stalls, as demonstrated in this paper. LSTM recurrent neural networks are also capable of perceiving long-range dependency as well as non-linearity in the temporal data and have been previously used by many researchers for solving a wide range of time-series related problems in aviation. Nanduri et al. demonstrated the application of LSTM recurrent neural networks in aircraft exceedance detection from multivariate time-series flight data~\cite{nanduri2016anomaly}. It has also been successfully used to predict excess vibration in aircraft engines, attitude (aircraft orientation) control and trajectory prediction~\cite{elsaid2016using, li2019intelligent, ma2020hybrid}.
This paper will present a detailed approach to utilize the temporal evolution of in-flight sensor data in conjunction with LSTM recurrent neural networks to identify stalls in advance. The flight characteristics of each aircraft is unique but the same procedure can be followed in order to train the model with actual sensor data from any particular aircraft to predict stalls in the respective aircraft. Moreover, the pre-trained weights can be used when re-training the model with sensor data from actual aircraft or certified full motion flight simulators, which will accelerate the training process.

\section{Aircraft stalls:}

\begin{figure}[ht]
  \centering
  \includegraphics[clip, trim=0.1cm 4cm 0.3cm 4cm, width=90mm,scale=1]{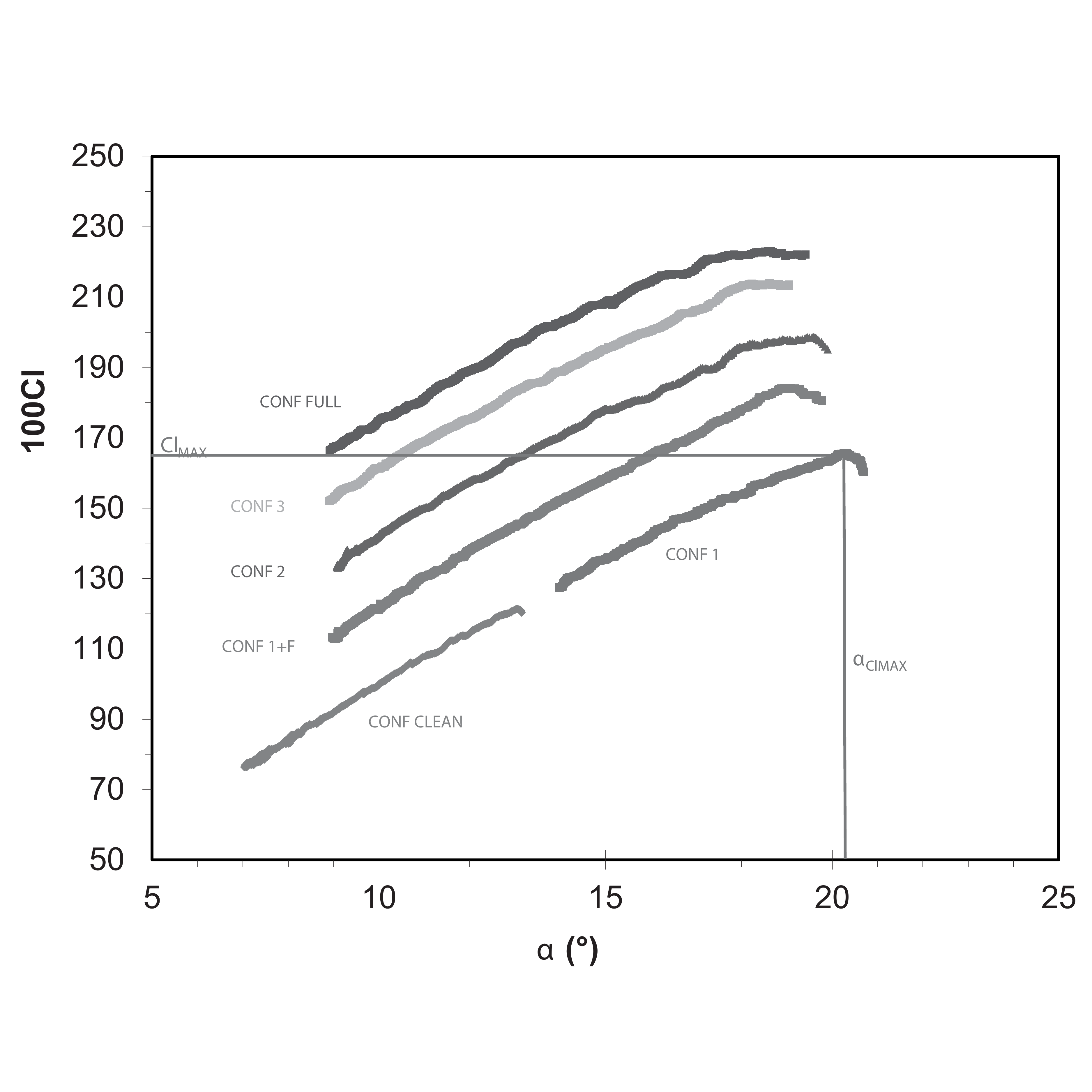}
  \caption{Lift coefficient vs angle of attack at different configurations [15]}
  \label{fig:fig1}
\end{figure}

Stall is defined as the point when maximum aerodynamic lift is achieved and a further increase in angle of attack $(\alpha)$ reduces lift~\cite{bolds2013stalling}. Reduction of lift during aircraft stalls are primarily caused by boundary layer separation on top of the wings. As the angle of attack $(\alpha)$ increases, adverse pressure gradient pushes the boundary layer backwards and this causes unsteady vortex which reduce lift drastically~\cite{chappell1968flow}. Loss of lift is often accompanied by diminished control response and high sink rate, which makes it extremely difficult to control the aircraft~\cite{stengel1982stalling}. However, the challenging aspect of stall is that, it can occur at any airspeed, at any attitude (aircraft orientation) and at any power setting~\cite{f2016a}. Various flight parameters directly influence the stall speed and its characteristics. Figure 1 shows the variation in the lift coefficient at various flight configurations (slat/flap position). Slats increase the angle of attack at which maximum lift is attained and flaps augment the magnitude of the maximum lift achievable. In addition to that, various combination of flight parameters such as thrust, airspeed, pitch, bank etc. affect the maximum lift coefficient and thus, are directly related to stall~\cite{bolds2013stalling}. For instance, on 18th September 2008, a Boeing 717-200 on final approach into Alice Springs experienced momentary stall warning. Investigation report revealed that the flight crew rolled the aircraft at a $30^\circ$ bank angle to intercept the final approach. A combination of increased wing loading, roll, pitch, low power, high air density at low altitude and changes to the relative wind led to the loss of airspeed as well as the increase in angle of attack~\cite{a2008a}. Thereby, resulting in 2 consecutive, momentary stick shaker activation, which is a form of stall warning. Investigation further revealed that the pilot in command (PIC) was severely fatigued which contributed to the diminished situational awareness. However, LSTM recurrent neural networks are capable of learning the unique temporal evolution of the various associated parameters prior to a stall and can predict potential stalls well in advance, which can assist the flight crew in avoiding flight conditions that may lead to stall warning or potential stalls. There are several regulations proposed by the Federal Aviation Administration (FAA) pertaining to aircraft stalls: 25·103 ‘Stall Speed’~\cite{BibEntry2020Nov(2)}, 25·203 ‘Stall Characteristics’~\cite{BibEntry2020Nov(3)}, 25·207 ‘Stall Warning ~\cite{BibEntry2020Nov(4)}. According to these laws, the aircraft stall warning should be triggered at a speed, $V_{sw}$, which exceeds the speed at which stall is identified by at least five knots or five percent calibrated airspeed.

\section{Methodology:}
The probability of stall at a future time step is dependent on a multivariate sequence of flight parameters with n historical timesteps. Therefore, the input can be characterized as a matrix of flight data and can be represented as: 
\begin{center}
\begin{equation}
X=
\begin{bmatrix}
P^{1}_{t-n} & P^{1}_{t-n+1} & P^{1}_{t-n+2} & . & . & . & P^{1}_{t-1} & P^{1}_{t}\\
P^{2}_{t-n} & P^{2}_{t-n+1} & P^{2}_{t-n+2} & . & . & . & P^{2}_{t-1} & P^{2}_{t}\\
P^{3}_{t-n} & P^{3}_{t-n+1} & P^{3}_{t-n+2} & . & . & . & P^{3}_{t-1} & P^{3}_{t}\\
. & . & . & . & . & . & . & .\\
. & . & . & . & . & . & . & .\\
. & . & . & . & . & . & . & .\\
P^{16}_{t-n} & P^{16}_{t-n+1} & P^{16}_{t-n+2} & . & . & . & P^{16}_{t-1} & P^{16}_{t}
\end{bmatrix}
\end{equation}
\end{center}

Here, the element $P^{16}_{t-n}$ represents the value of the 16th flight parameter (aircraft vertical speed), at the $(t-n)^{th}$ timestep. The matrix can be simplified and expressed as an input vector with each element representing a vector of flight parameters.

\begin{center}
\begin{equation}
X=
\begin{bmatrix}
x_{t-n}, x_{t-n+1}, x_{t-n+2} & . & . & . & x_{t}
\end{bmatrix}
\end{equation}
\end{center}

where

\begin{center}
\begin{equation}
x_t=
\begin{bmatrix}
P^{1}_{t}\\
P^{2}_{t}\\
P^{3}_{t}\\
.\\
.\\
.\\
P^{16}_{t}
\end{bmatrix}
\end{equation}
\end{center}

Each input vector X is tagged with a label of 1 or 0, representing the stall warning state at a future timestep (t+n). The input vector is then passed into an LSTM layer consisting of a sequence of LSTM cells corresponding to each timestep. For any given timestep t, the LSTM cell consists of three gate units, input gate $i_t$ , forget gate $f_t$ and output gate $o_t$ ~\cite{hochreiter1997long}. The LSTM cell takes the flight parameter vector $x_t$ at time step t and the output of the previous LSTM cell $h_(t-1)$ corresponding to the $(t-1)^{th}$ timestep as inputs. In addition, it also takes the previous cell state $C_{(t-1)}$ as input, which facilitates the transport of relevant information throughout the sequence. The updated cell state, $C_t$ and the cell output $h_t$ are passed onto the next cell to repeat the process throughout the sequence and the weight matrices are updated in order to minimize the loss function.

\section{Data collection and preprocessing:}
The training and test data were collected from the commercially available flight simulator- X-plane 11. Apart from pilot training, X-plane is also extensively used for research and as an engineering tool by researchers, defense contractors, air forces, aircraft manufacturers, Cessna as well as NASA for applications ranging from flight training to concept design and flight testing ~\cite{agha2017system, bittar2013a, meyer2016a, etin-a, junior-a, bittar-a, akyurek-a, bittar2014a, ackerman2014a}. Nanduri et al. demonstrated the capability of X-plane in generating large volumes of Flight Operations Quality Assurance (FOQA) data which can be used for data mining algorithms and machine learning ~\cite{nanduri-b}. X-plane implements the blade element theory in its flight dynamics modules (FDM) to estimate the forces acting on an

\begin{figure}[ht]
  \centering
  \includegraphics[clip, trim=0.1cm 4cm 0.3cm 3cm, width=130mm,scale=1]{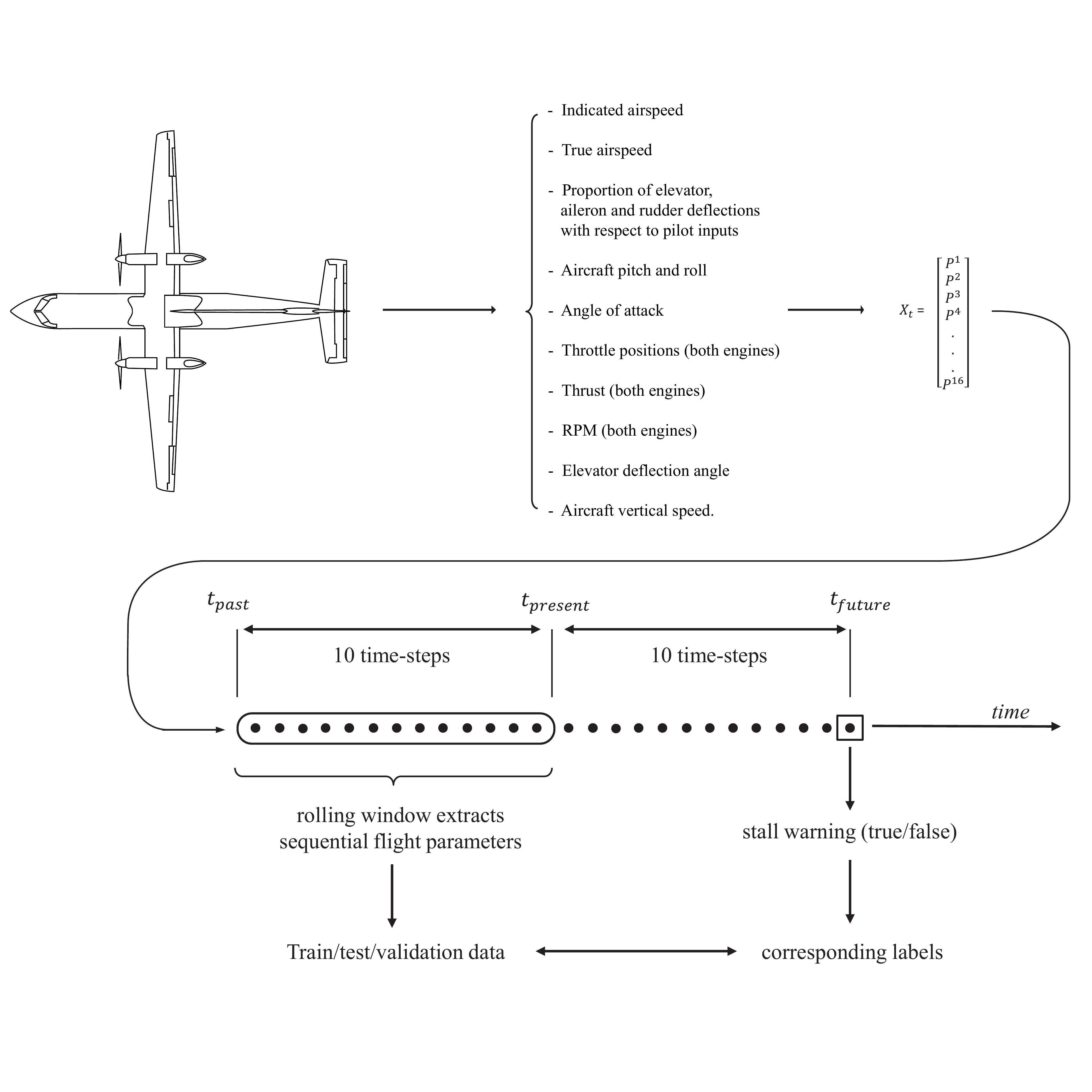}
  \caption{Schematic representation of data formatting and training data preparation}
  \label{fig:fig2}
\end{figure}

aircraft and closely model the aircraft’s flight characteristics ~\cite{xplane-b}. In this method, the aircraft lifting surfaces are broken down into several longitudinal strips and the forces calculated in each strip are integrated to calculate the total lifting force. Thus, X-plane is able to demonstrate high degree of accuracy in modeling and simulating stall characteristics to within 4.67\% of the actual aircraft ~\cite{thong2011a}. Different flight parameters are exported from X-plane as temporal data. Each time step is associated with a flight parameter vector $x_t$. The exported data also represents the aircraft’s stall warning state as a boolean data type (1 when stall warning=true and 0 when stall warning=false). This was used to identify stall warnings in the sequence data and collect the flight parameter sequence prior to stall warnings. Each sequence is then labeled with the stall warning state at a future times-step in order to prepare the final dataset. Figure 2 provides a schematic of the data formatting procedure and preparation of train, test and validation dataset. Since the dataset is composed of parameters with highly varying magnitudes, range and units, standardization is used to bring all the features to the same scale. This results in a distribution with mean of 0 and a standard deviation of 1.

\begin{center}
\begin{equation}
scaled\ feature\ x^{'} =\frac{x-\overline{x}}{\sigma}
\end{equation}
\end{center}

Where, $\overline{x}$ denotes the mean and $\sigma$ denotes the standard deviation of the dataset.

\section{Dataset description:}
Around 53000 seconds of flight data were collected from the ‘Bombardier DHC-8-400 (Q400)’ turbo-prop aircraft in the simulator, then processed to create the final data-set, containing aggregated 26400 seconds of multivariate time series with 16 flight parameters:  indicated airspeed, true airspeed, proportion of elevator, aileron and rudder deflections with respect to pilot inputs, aircraft pitch and roll, angle of attack, throttle positions (both engines), thrust (both engines), RPM (both engines), elevator deflection angle and aircraft vertical speed. The flight data is primarily collected from the climb, cruise, maneuver and descent phase since it contributes to around 80\% of the flight time and highest number of stall incidents have occurred at this particular phase of the flight (Figure 3) ~\cite{BibEntry2020Nov}. Other phases of flight have not been

\begin{figure}[ht]
  \centering
  \includegraphics[clip, trim=10cm 6cm 7cm 6cm, width=120mm,scale=1]{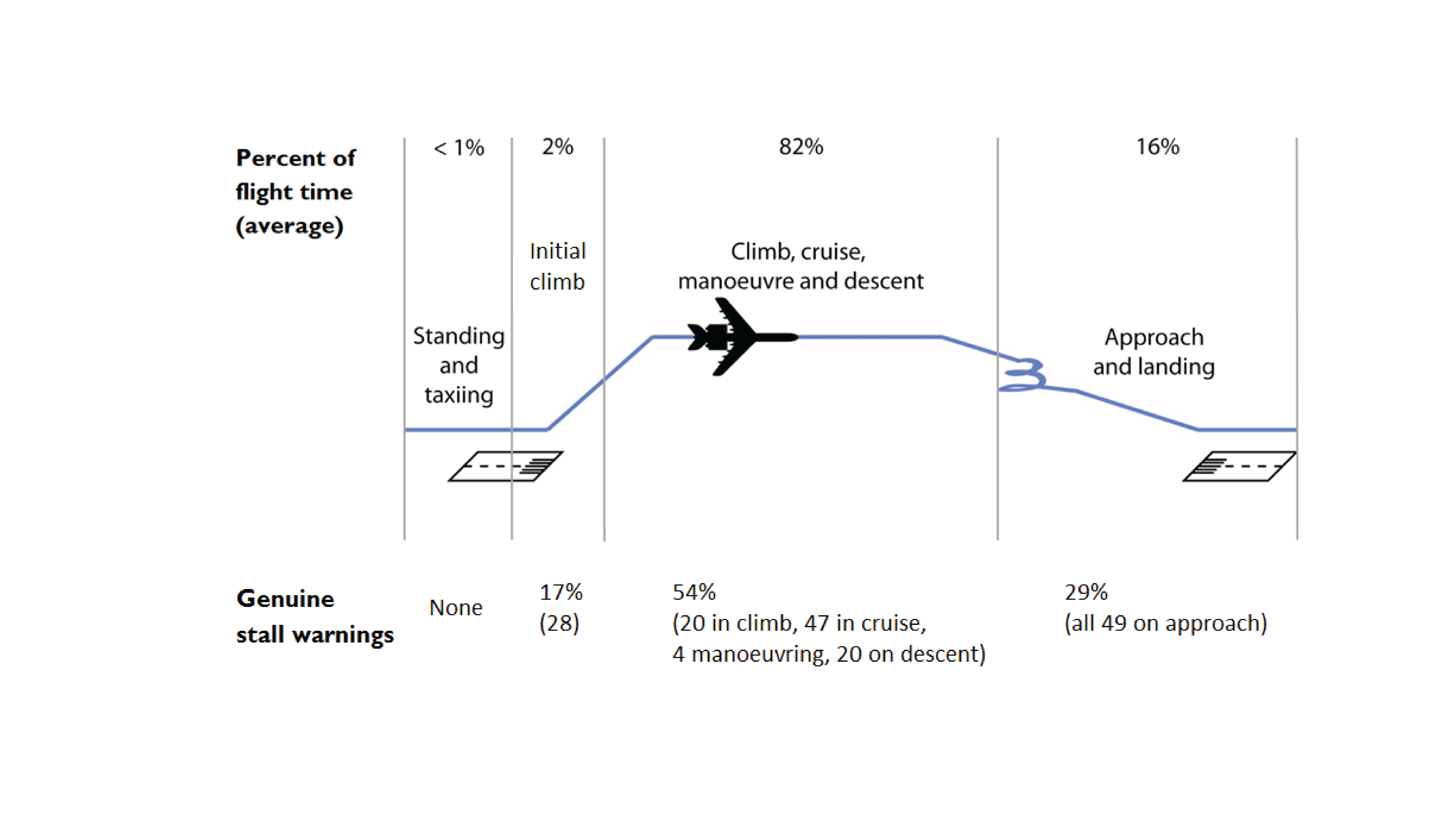}
  \caption{Stall warning incidents at different stages of flight according to ATSB ~\cite{BibEntry2020Nov} }
  \label{fig:fig3}
\end{figure}

considered in this study, since the primary aim is to demonstrate the potential of artificial intelligence in predicting stall. Nevertheless, the risk of stall is also high during approach, landing and go-arounds as well and therefore are considered for inclusion in future studies. The training dataset consists of 2040 (1020 stall positive and 1020 stall negative) samples while the validation and test dataset both consist of 300 (150 stall positive and 150 stall negative) samples each. Therefore, the train, test and validation datasets are well balanced. Each sample in the dataset are comprised of 16 flight parameters at 10 previous timesteps. 

\section{Neural network architecture and Bayesian hyper-parameter optimization:}

Three LSTM recurrent neural network architectures (“architecture A”, “architecture B” and “architecture C”) were designed to identify impending stall 10 seconds in advance. The structure of the three architectures are provided in Table 1.

\begin{figure}[ht]
  \centering
  \includegraphics[clip, trim= 2.5cm 15.5cm 2.5cm 2.5cm, width=150mm,scale=1]{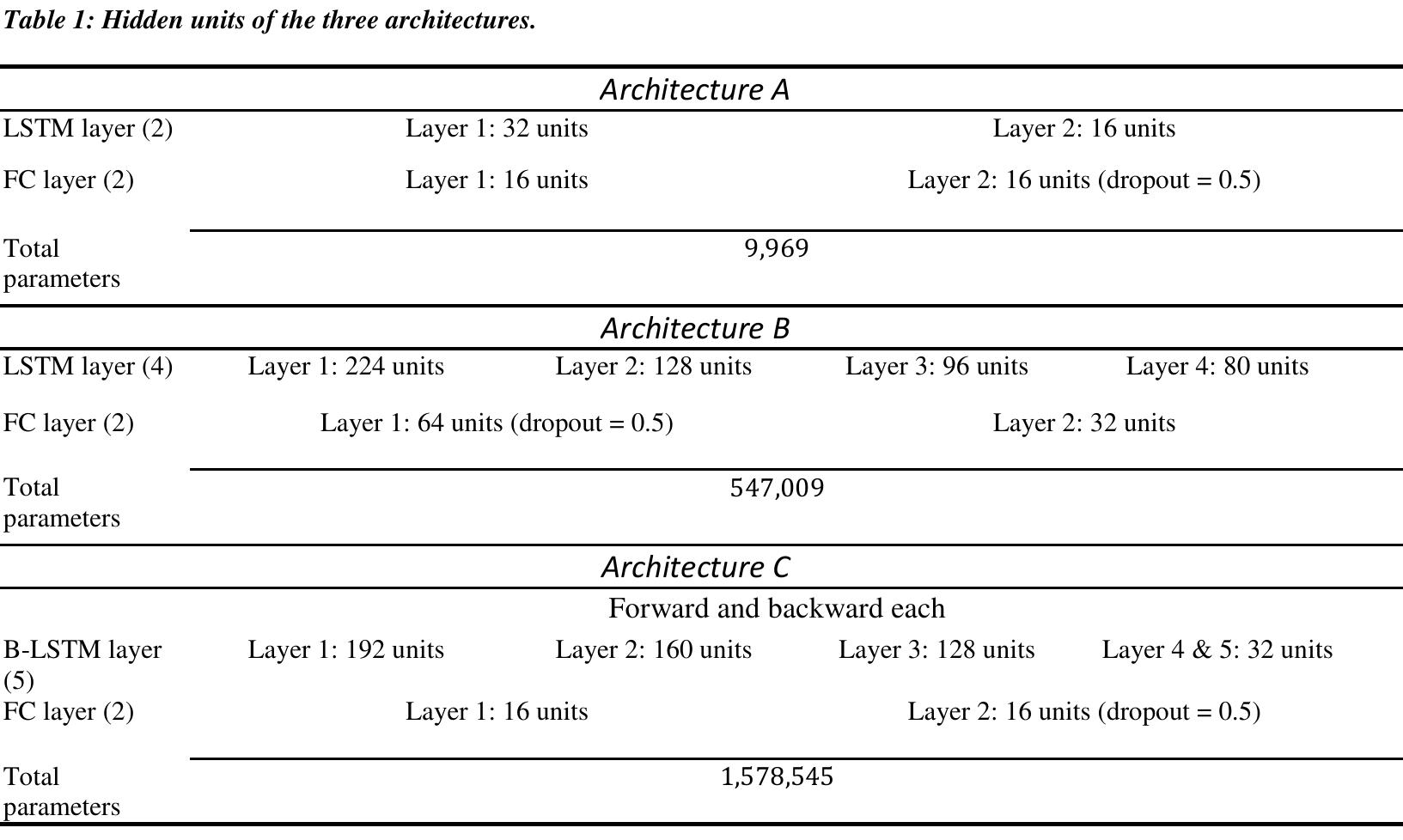}
  \label{fig:fig4}
\end{figure}

 All architectures take flight parameter input from the previous ten timesteps including the current timestep t. The general structure of the three architectures are shown in Figure 4. “Architecture A” has 2 recurrent LSTM layers and the output from the final LSTM layer is fed into fully connected layers. Each LSTM layer consists of 10 cells corresponding to 10 previous timesteps respectively. Therefore, the input matrix dimension for each training example is 16×10 (16 flight parameters and 10 timesteps). The 2 LSTM layers have 32 and 16 hidden units respectively followed by 2 fully connected layers with 16 hidden units each. The last fully connected layer with dropout of 0.5, passes information to the final output unit which uses a Boolean data type to represent the possibility of a stall from the temporal input data. Dropout is only applied to the non-recurrent layers based on the suggestions by Pham et al. and Zaremba et al  ~\cite{pham2014a,zaremba2014a}.  Each of the 10 LSTM cells have 4 trainable weights and 4 trainable biases which results in a total of 9,969 trainable parameters in “Architecture A”. 

\begin{figure}[ht]
  \centering
  \includegraphics[clip, trim= 0.5cm 1cm 0cm 1.5cm, width=160mm,scale=1]{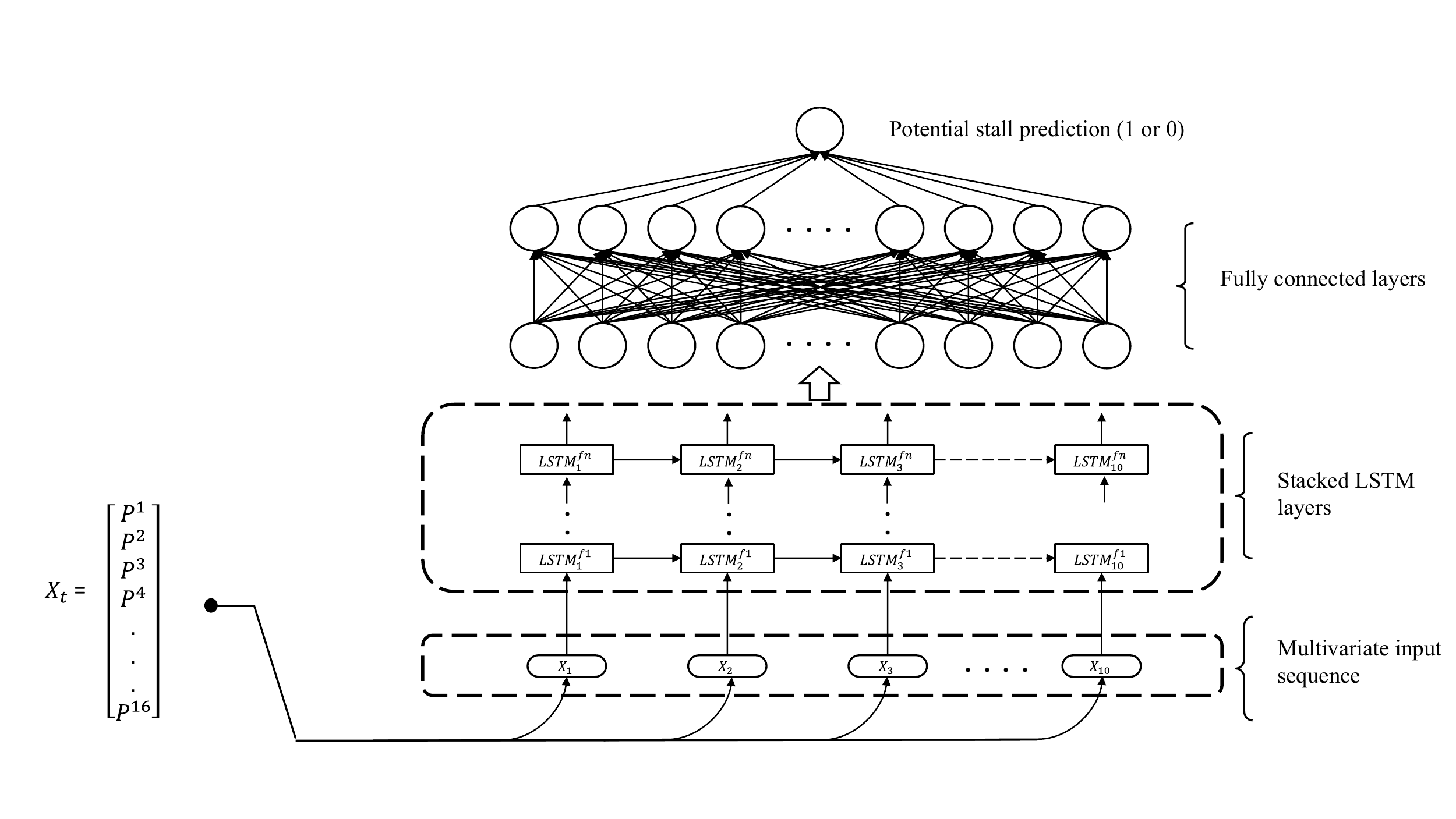}
  \caption{Neural network architecture}
  \label{fig:fig5}
\end{figure}

The hyperparameters of “Architecture B” and “Architecture C” on the contrary have been optimized using Bayesian Optimization with Gaussian Process and is highlighted in Figure 5. Bayesian optimization is a sequential model based optimization (SMBO) algorithm, which uses surrogate function as a proxy to make informed decision and find the extrema of the objective function ~\cite{brochu2010a}. It utilizes the “Bayes rule”, which in this case seeks to find the probability of accuracy, given a particular combination of hyper-parameters. Mathematically it can be defined as:

\begin{center}
\begin{equation}
P(Accuracy|Hyperparameter\ choice) =\frac{P(Hyperparameter\ choice|Accuracy)P(Accuracy)}{P(Hyperparameter\ choice)}
\end{equation}
\end{center}

Initially, a probabilistic or surrogate model is constructed by evaluating certain arbitrary combinations of hyperparameters. The surrogate model, which is an approximation of the objective function is created using the gaussian process ~\cite{rasmussen2003a}. The gaussian process is able to capture an effective estimate of the objective function after sufficient iterations. The subsequent hyperparameters used in the optimization process are chosen based on their performance on this surrogate model, which is characterized by the “Expected Improvement”, also known as the acquisition function. The “Expected Improvement” function, $EI_{y^{\ast}}(x)$ introduces a threshold value $y^\ast$ and searches for a combination of hyperparameters x using the surrogate model P(y|x), such that the accuracy y exceeds the threshold value. It is mathematically represented as:

\begin{center}
\begin{equation}
EI_{y^{\ast}}(x) =\int_{-\infty}^{y^{\ast}} (y^{\ast}-y)P(y|x)dy
\end{equation}
\end{center}

This function assists the optimizer in deciding the best candidate hyperparameters x to evaluate next. The chosen combination of hyperparameter is then used to train and evaluate the neural network’s accuracy on the validation dataset. 

\begin{figure}[ht]
  \centering
  \includegraphics[clip, trim= 0cm 0cm 0cm 0cm, width=160mm,scale=1]{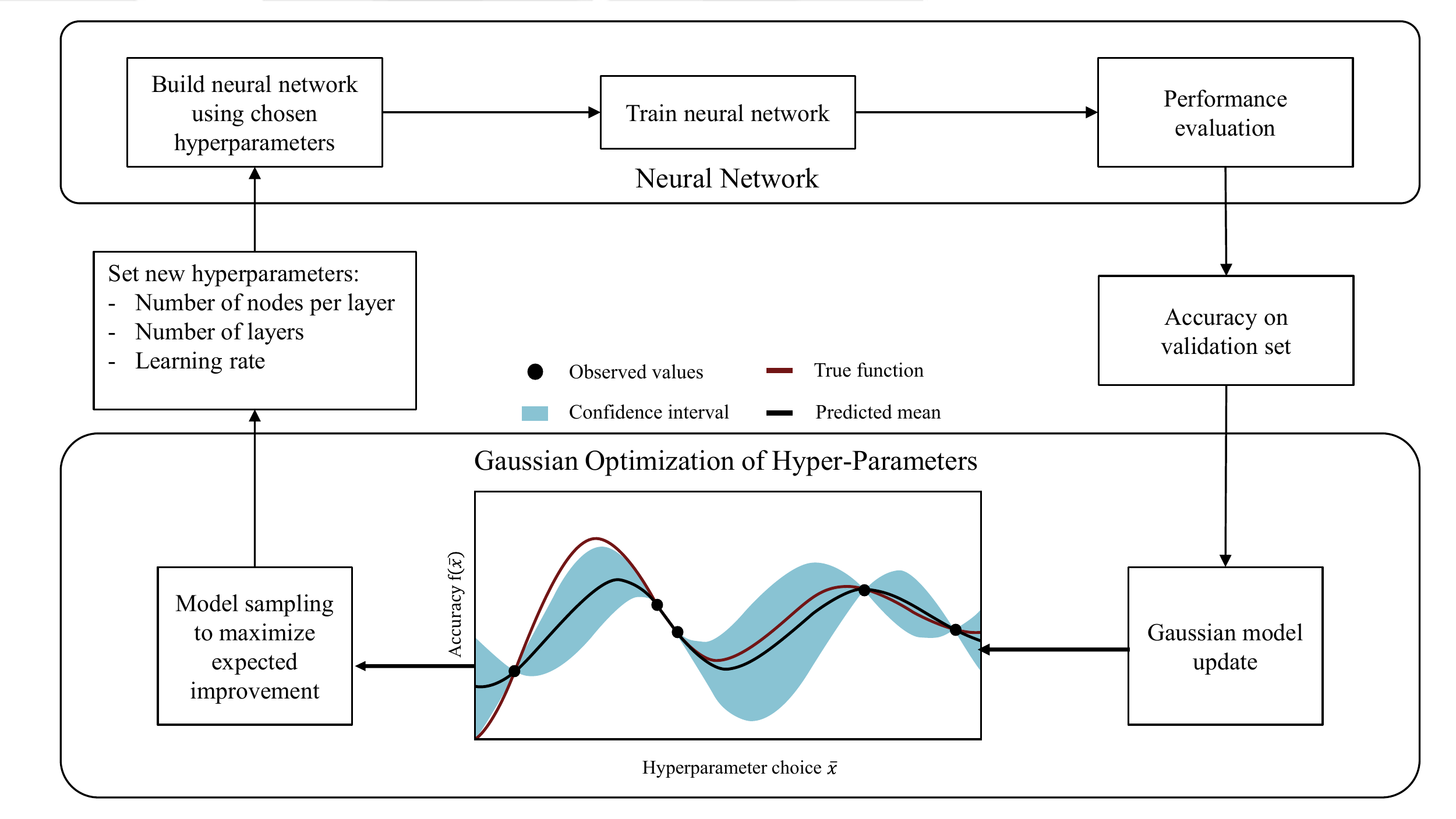}
  \caption{Bayesian hyperparameter optimization with Gaussian process}
  \label{fig:fig6}
\end{figure}

The accuracy metric for the given hyperparameter combination is used to update the surrogate model and the process is repeated for a fixed number of iterations to seek the set of hyperparameters corresponding to the maxima of the objective function or in this case the accuracy.

“Architecture B” optimized by Bayesian optimization, consists of four recurrent unidirectional LSTM layers followed by fully connected layers. The LSTM layers have 224, 128, 96, 80 hidden units respectively and the fully connected layers have 64 and 32 hidden units respectively. A dropout of 0.5 is applied to the first fully connected layer with 64 hidden units.  
“Architecture C” is also tuned using Bayesian optimization similar to “Architecture B”. However, instead of unidirectional LSTM layers, it consists of 5 bidirectional stacked LSTM layers with 192, 160, 128, 32 and 32 hidden units in both directions. The outputs of the bidirectional LSTM layers are passed into 2 fully connected layers with 16 hidden units. In addition, a dropout of 0.5 is applied to the 2nd fully connected layer.

\section{Experimental setup and training:}
All models are trained on 2040 training data and tested on 300 separate test data. Furthermore, a different set of 300 data are used as validation set to tune the models. Each model is trained for 100 epochs with Adam (Adaptive moment estimation) optimizer ~\cite{kingma2014a}. To prevent over-fitting, dropout of 0.5 is added to the non-recurrent layer. The learning rate for “Architecture B” and “Architecture C” are chosen using Bayesian hyperparameter optimization. “He” initialization is used to initialize the weights and prevent activation outputs from vanishing or exploding during forward propagation. Since, we are dealing with a binary classification problem, binary cross-entropy or log loss is used as the cost function and is expressed as follows:

\begin{center}
\begin{equation}
J =-\frac{1}{N}\sum _{i=1}^{N} y_i\cdot log(p(y_i))+(1-y_i)\cdot log(1-p(y_i))
\end{equation}
\end{center}

Finally, in order to evaluate the performance of each models, the training and validation loss are recorded and plotted after each epoch. The hyperparameter choice for each model is presented in Table 2

\begin{figure}[ht]
  \centering
  \includegraphics[clip, trim= 2.5cm 19.7cm 2.5cm 2.5cm, width=150mm,scale=1]{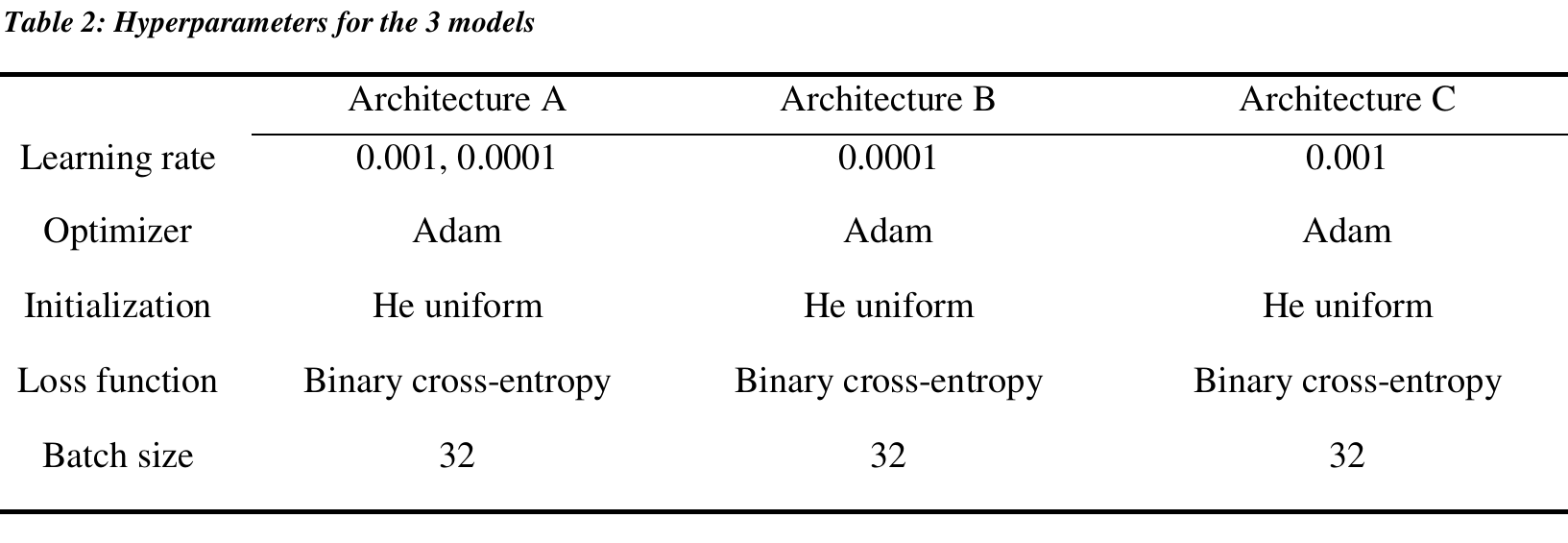}
  \label{fig:fig7}
\end{figure}

\section{Results and analysis:}
In order to evaluate the performance, we report the accuracy, Area Under the ROC Curve (AUC), precision, recall and F score of the three models. These results, representing the performance of the 3 models on the test dataset, are presented in Table 3.

\begin{figure}[ht]
  \centering
  \includegraphics[clip, trim= 2.5cm 19cm 2.5cm 2.5cm, width=150mm,scale=1]{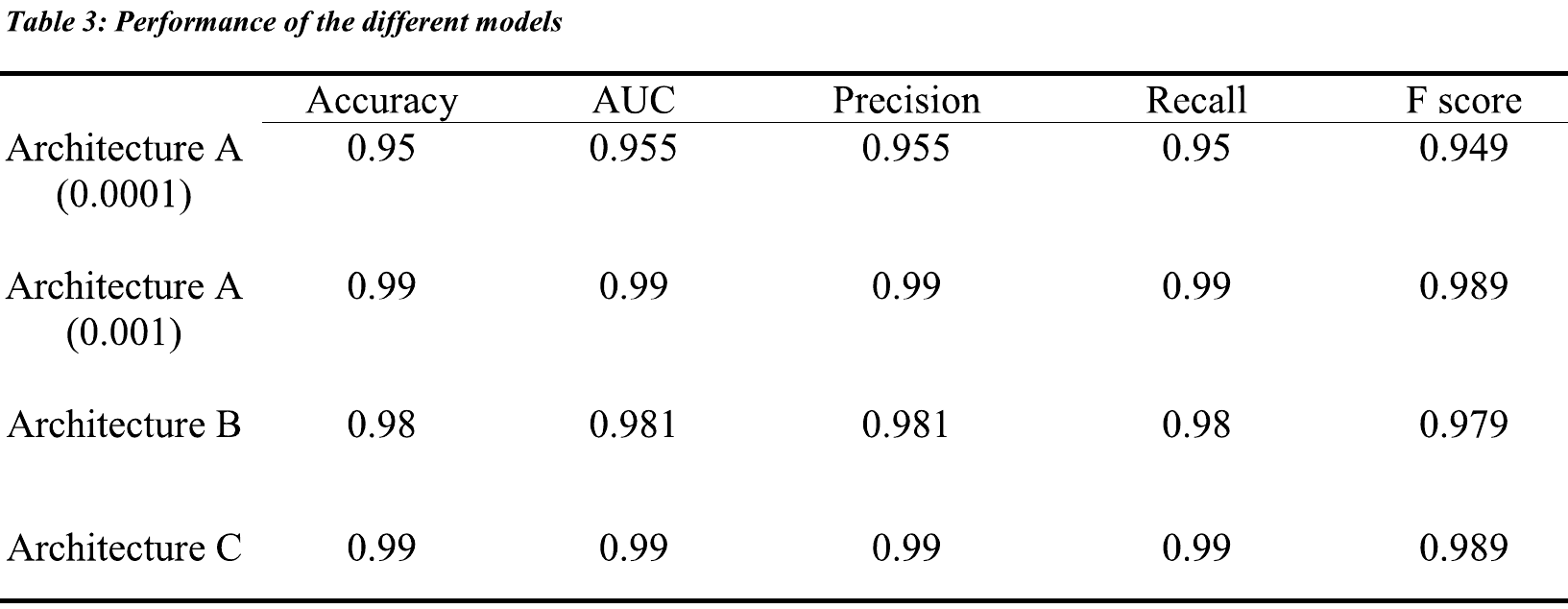}
  \label{fig:fig8}
\end{figure}

The training and validation loss at each epoch for the 3 models reveal that “Architecture B” and “Architecture C” are able to minimize loss at fewer epochs compared to “Architecture A” (Figure 6).

\begin{figure}[ht]
  \centering
  \includegraphics[clip, trim= 3cm 3cm 3cm 3cm, width=85mm,scale=1]{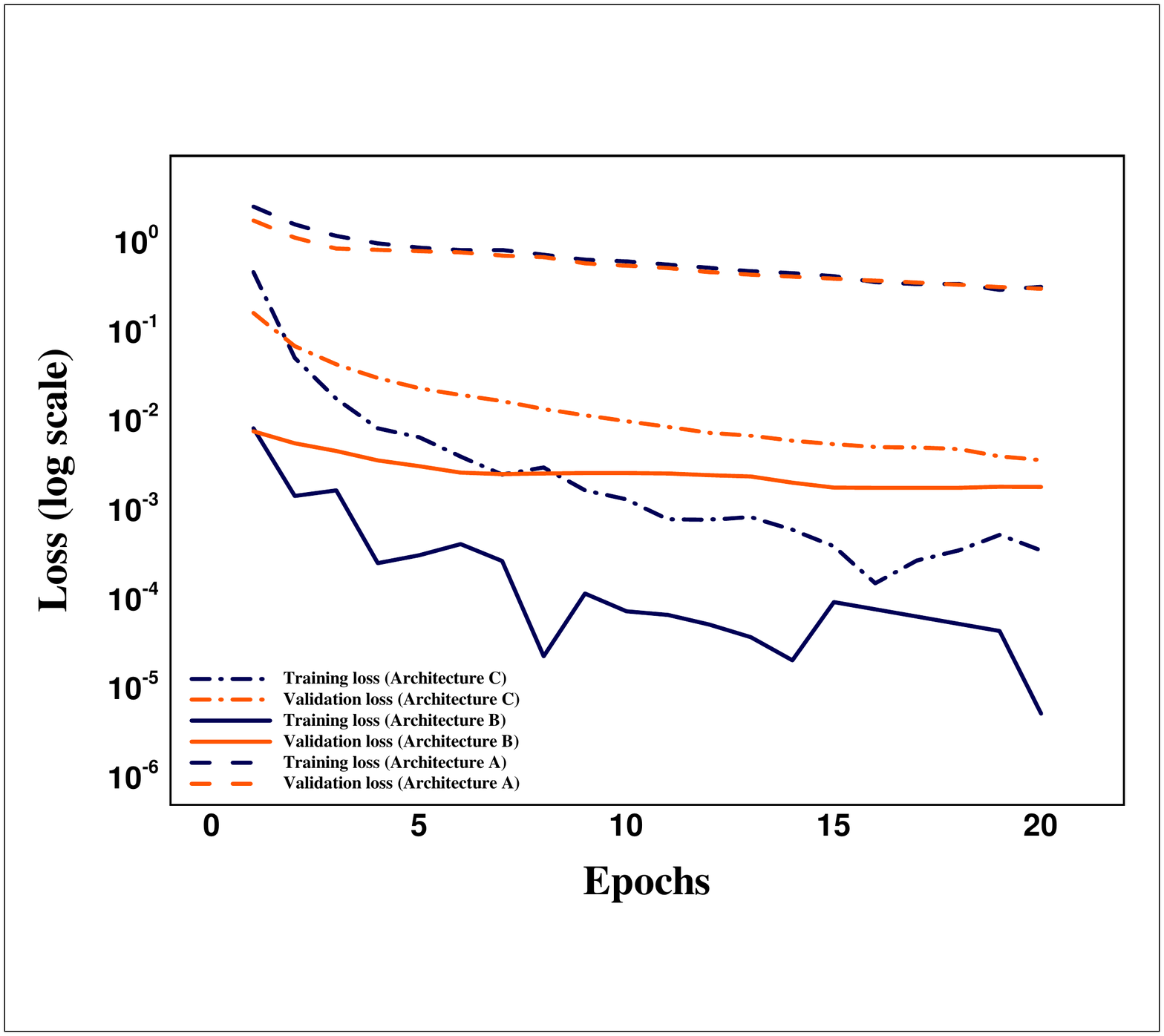}
  \caption{Loss at each epoch for different models}
  \label{fig:fig9}
\end{figure}

Moreover, as seen from table 3, the accuracy of “Architecture B” and “Architecture C”, exceeds that of “Architecture A”. However, if a learning rate (LR) of 0.001 is used instead of 0.0001, “Architecture A” can achieve similar performance as the other two at fewer epochs.To further demonstrate the effectiveness of the models, the breakdown of prediction accuracy for the different models has been presented in Table 4. 

\begin{figure}[ht]
  \centering
  \includegraphics[clip, trim= 2.5cm 17.5cm 2.5cm 2.5cm, width=150mm,scale=1]{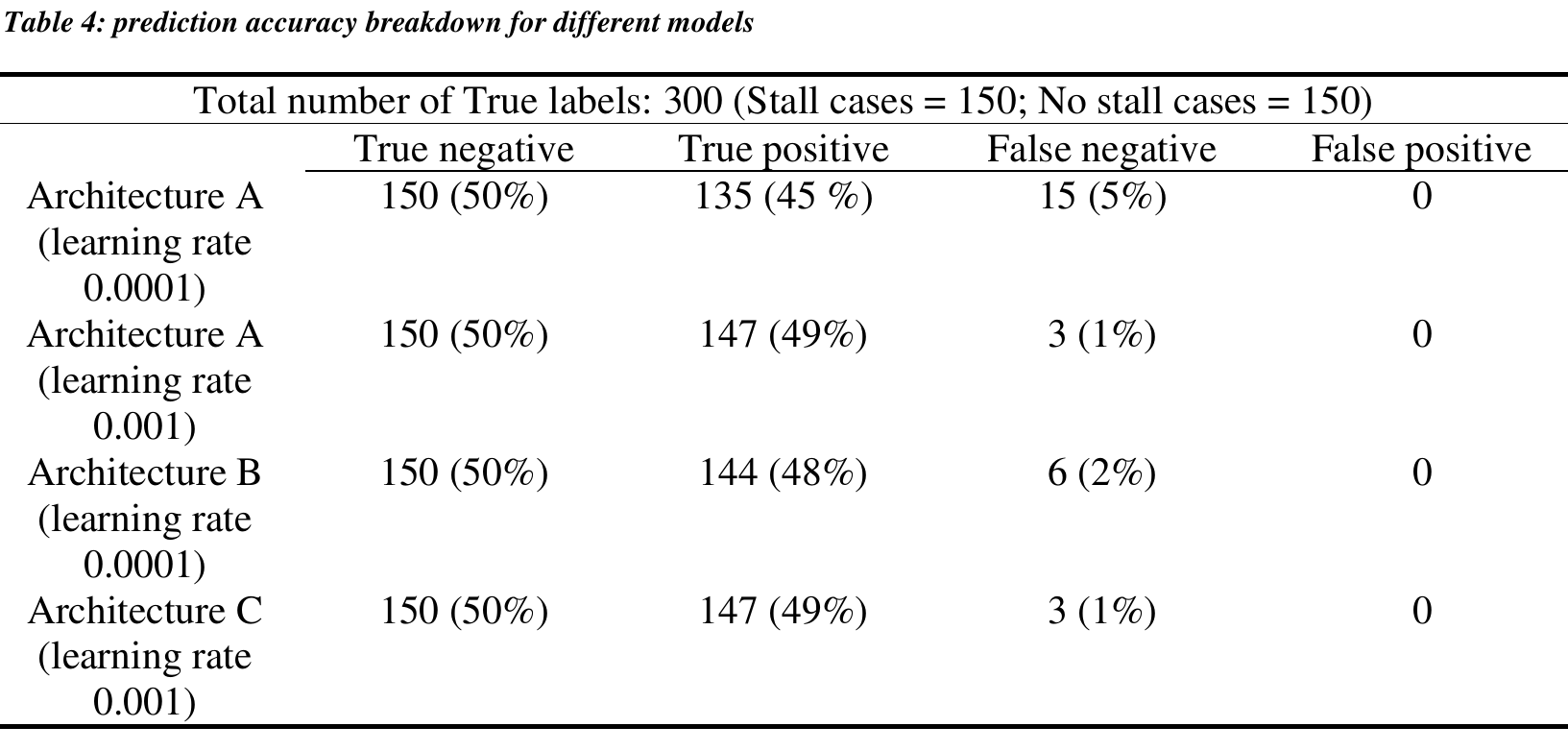}
  \label{fig:fig10}
\end{figure}

“Architecture A” trained with a learning rate of 0.0001, has a 5\% false negative rate with 15 misclassified samples but is able to accurately identify all the “no-stall” cases. Out of 300 samples in the test set, “Architecture B” and “Architecture C” are able to correctly identify 294 and 297 sample respectively. Upon, close inspection of the three miss classified cases by “Architecture C”, it is seen that these are secondary stalls induced by abrupt control inputs. Figure 7 demonstrates the differences between a misclassified and correctly classified sample. The misclassified example (abrupt stall) has insufficient variation in the temporal data to correctly identify a potential stall, whereas the correctly classified sample (gradual stall) exhibits temporal variations in the flight parameters leading up to the stall. “Architecture A (LR=0.0001)” and “Architecture B” exhibit >95\% accuracy whereas “Architecture C” exhibits 100\% accuracy in predicting gradual stalls on the validation set. Therefore, the deep learning-based approach can provide useful warning to the pilots in case of a gradual stall.

\begin{figure}[ht]
  \centering
  \includegraphics[clip, trim= 0.5cm 5.5cm 0.5cm 5cm, width=150mm,scale=1]{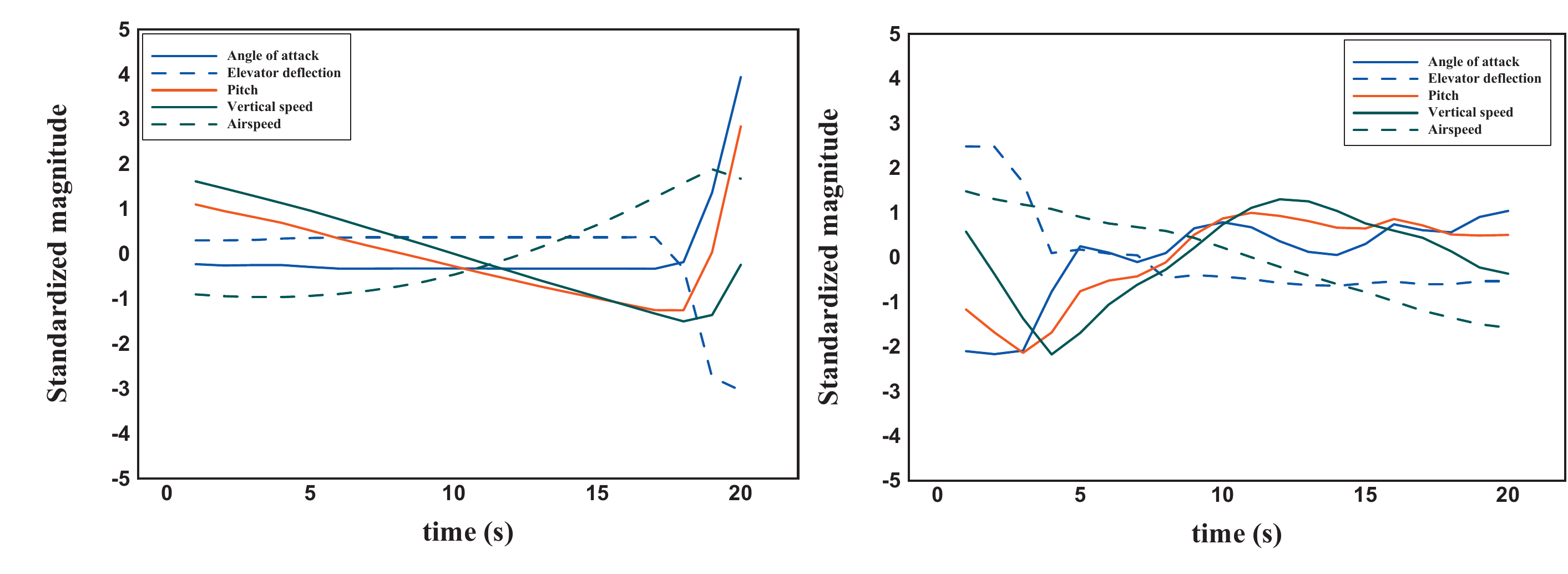}
  \caption{Comparison of a misclassified sample (left) with a correctly classified sample (right)}
  \label{fig:fig11}
\end{figure}

\section{Limitations and future scopes:}
The deep learning approach presented in this paper is able to predict gradual stalls 10 seconds in advance. However, this deep learning method is unable to predict secondary stalls or stalls arising from sudden maneuvers (figure 7). A secondary stall occurs after recovery from a preceding stall and is typically caused by abrupt control inputs or when the pilot attempts to return to the desired flightpath too quickly thus exceeding the critical angle of attack a second time. Furthermore, unpredictable events such as sudden gusts, turbulence or extreme maneuvers can rapidly push the aircraft beyond its flight envelope. As a result, there isn’t sufficient temporal information during a secondary stall or an abrupt stall for the LSTM recurrent neural network to make an accurate prediction. Nevertheless, the AI assisted stall prediction approach can complement conventional stall warning systems in predicting gradually developing stall and add an additional layer of safety to improve the flight crew’s situational awareness.

The current study focuses on stall during cruise phase of the flight. To allow the neural network to accurately predict stall during other phases of flight, additional parameters need to be taken into consideration. These parameters may include landing gear state, flap and slat positions, aircraft weight and center of gravity, altitude (to account for density and other atmospheric parameters) or even icing on the wings. All of these parameters, change during the course of a flight and can alter the stall characteristics of the aircraft. In addition, since the flight dynamics of each aircraft is unique, a general model is not applicable and the model needs to be trained separately for individual aircrafts. Though modern simulators are highly accurate in capturing the flight dynamics at various stages of flight, it is difficult to rely on simulation solely when it comes to flight safety. Therefore, flight test engineers and test pilots conduct simulation as well as test flights for certification of new aircrafts ~\cite{bolds2013stalling}. These test flights provide ample opportunity to gather useful flight data from the actual aircraft under controlled environment. Therefore, these test flight data can be used to train the neural network for better accuracy on actual flights.

\section{Conclusion:}
This paper presents a deep learning-based approach to predict potential stalls during flight. Uni-directional and bi-directional, stacked LSTM recurrent neural network architectures have been proposed to identify impending stalls from the in-flight sensor data. The models analyze various flight parameters from the previous timesteps to predict potential stalls in future timesteps. The models were evaluated on a test dataset of previously unseen flight parameter samples and were able to accurately predict impending stalls from these samples. The multilayer bi-directional LSTM architecture exhibits slightly better performance in terms of accuracy; however, all the models are capable of predicting potential stalls with high degree of accuracy. The limitations of the existing model and prospects of future study have also been addressed. LSTM neural networks have great potential to predict stalls or any dangerous flight events and thus augment the flight crew’s situational awareness. Therefore, this deep learning approach can be further studied to include stall prediction in all phases of flight.

\bibliographystyle{IEEEtran}
\bibliography{Bibliography}

\begin{thebibliography}{10}
\providecommand{\url}[1]{#1}
\csname url@samestyle\endcsname
\providecommand{\newblock}{\relax}
\providecommand{\bibinfo}[2]{#2}
\providecommand{\BIBentrySTDinterwordspacing}{\spaceskip=0pt\relax}
\providecommand{\BIBentryALTinterwordstretchfactor}{4}
\providecommand{\BIBentryALTinterwordspacing}{\spaceskip=\fontdimen2\font plus
\BIBentryALTinterwordstretchfactor\fontdimen3\font minus
  \fontdimen4\font\relax}
\providecommand{\BIBforeignlanguage}[2]{{%
\expandafter\ifx\csname l@#1\endcsname\relax
\typeout{** WARNING: IEEEtran.bst: No hyphenation pattern has been}%
\typeout{** loaded for the language `#1'. Using the pattern for}%
\typeout{** the default language instead.}%
\else
\language=\csname l@#1\endcsname
\fi
#2}}
\providecommand{\BIBdecl}{\relax}
\BIBdecl

\bibitem{airplanes2016statistical}
\BIBentryALTinterwordspacing
``Statistical summary of commercial jet airplane accidents:worldwide operations
  1959--2018,'' Boeing Commercial Airplanes, Seattle, WA 98124-2207, Tech. Rep.
  50th edition, Sep. 2019. [Online]. Available:
  \url{http://www.boeing.com/resources/boeingdotcom/company/about_bca/pdf/statsum.pdf}
\BIBentrySTDinterwordspacing

\bibitem{collins2015stall}
\BIBentryALTinterwordspacing
D.~J. Kenny, J.~Collins, and A.~Sable, ``Stall and spin accidents: Keep the
  wings flying,'' AOPA Air Safety Institute, 421 Aviation Way, Frederick, MD
  21701, Tech. Rep., 2015. [Online]. Available:
  \url{https://www.aopa.org/-/media/files/aopa/home/pilot-resources/safety-and-proficiency/accident-analysis/special-reports/stall_spin.pdf}
\BIBentrySTDinterwordspacing

\bibitem{aircraft201423rd}
``23rd joseph t. nall report: General aviation accidents in 2011,'' AOPA Air
  Safety Institute, 421 Aviation Way, Frederick, MD 21701, Tech. Rep., 2014.

\bibitem{BibEntry2020Nov}
\BIBentryALTinterwordspacing
``Stall warnings in high capacity aircraft: The australian context 2008 to
  2012,'' Australian Transport Safety Bureau, S62 Northbourne Avenue Canberra,
  Australian Capital Territory 2601, Tech. Rep., Nov. 2013. [Online].
  Available: \url{https://www.atsb.gov.au/publications/2012/ar-2012-172}
\BIBentrySTDinterwordspacing

\bibitem{f2016a}
\emph{Airplane Flying Handbook- Maintaining Aircraft Control: Upset Prevention
  and Recovery Training}.\hskip 1em plus 0.5em minus 0.4em\relax P.O. Box
  25082, Oklahoma City, OK 73125: FEDERAL AVIATION ADMINISTRATION, 2016, ch.~4.

\bibitem{hochreiter1997long}
S.~Hochreiter and J.~Schmidhuber, ``Long short-term memory,'' \emph{Neural
  computation}, vol.~9, no.~8, pp. 1735--1780, 1997.

\bibitem{dabek2015neural}
F.~Dabek and J.~J. Caban, ``A neural network based model for predicting
  psychological conditions,'' in \emph{International conference on brain
  informatics and health}.\hskip 1em plus 0.5em minus 0.4em\relax Springer,
  2015, pp. 252--261.

\bibitem{lipton2015learning}
Z.~C. Lipton, D.~C. Kale, C.~Elkan, and R.~Wetzel, ``Learning to diagnose with
  lstm recurrent neural networks,'' \emph{arXiv preprint arXiv:1511.03677},
  2015.

\bibitem{zhao2017learning}
R.~Zhao, R.~Yan, J.~Wang, and K.~Mao, ``Learning to monitor machine health with
  convolutional bi-directional lstm networks,'' \emph{Sensors}, vol.~17, no.~2,
  p. 273, 2017.

\bibitem{jiao2018short}
R.~Jiao, T.~Zhang, Y.~Jiang, and H.~He, ``Short-term non-residential load
  forecasting based on multiple sequences lstm recurrent neural network,''
  \emph{IEEE Access}, vol.~6, pp. 59\,438--59\,448, 2018.

\bibitem{nanduri2016anomaly}
A.~Nanduri and L.~Sherry, ``Anomaly detection in aircraft data using recurrent
  neural networks (rnn),'' in \emph{2016 Integrated Communications Navigation
  and Surveillance (ICNS)}.\hskip 1em plus 0.5em minus 0.4em\relax Ieee, 2016,
  pp. 5C2--1.

\bibitem{elsaid2016using}
A.~ElSaid, B.~Wild, J.~Higgins, and T.~Desell, ``Using lstm recurrent neural
  networks to predict excess vibration events in aircraft engines,'' in
  \emph{2016 IEEE 12th International Conference on e-Science
  (e-Science)}.\hskip 1em plus 0.5em minus 0.4em\relax IEEE, 2016, pp.
  260--269.

\bibitem{li2019intelligent}
B.~Li, P.~Gao, X.~Li, and D.~Chen, ``Intelligent attitude control of aircraft
  based on lstm,'' in \emph{IOP Conference Series: Materials Science and
  Engineering}, vol. 646.\hskip 1em plus 0.5em minus 0.4em\relax IOP
  Publishing, 2019, p. 012013.

\bibitem{ma2020hybrid}
L.~Ma and S.~Tian, ``A hybrid cnn-lstm model for aircraft 4d trajectory
  prediction,'' \emph{IEEE Access}, vol.~8, pp. 134\,668--134\,680, 2020.

\bibitem{bolds2013stalling}
P.~J. Bolds-Moorehead, V.~Chaney, T.~Lutz, and S.~Vaux, ``Stalling transport
  aircraft,'' \emph{The Aeronautical Journal}, vol. 117, no. 1198, pp.
  1183--1206, 2013.

\bibitem{chappell1968flow}
P.~Chappell, ``Flow separation and stall characteristics of plane,
  constant-section wings in subcritical flow,'' \emph{The Aeronautical
  Journal}, vol.~72, no. 685, pp. 82--90, 1968.

\bibitem{stengel1982stalling}
R.~F. Stengel and W.~B. Nixon, ``Stalling characteristics of a general aviation
  aircraft,'' \emph{Journal of Aircraft}, vol.~19, no.~6, pp. 425--434, 1982.

\bibitem{a2008a}
\BIBentryALTinterwordspacing
``Investigation: Ao-2008-064 - stickshaker activation – boeing 717-200,''
  Australian Transport Safety Bureau, 62 Northbourne Ave, Canberra City,
  Australian Capital Territory, 2601, Tech. Rep., 2011. [Online]. Available:
  \url{https://www.atsb.gov.au/publications/investigation_reports/2008/aair/ao-2008-064/}
\BIBentrySTDinterwordspacing

\bibitem{BibEntry2020Nov(2)}
``14 cfr {\ifmmode\S\else\textsection\fi} 25.103 - stall speed.'' LII / Legal
  Information Institute,https://www.law.cornell.edu/cfr/text/14/25.103, Federal
  Aviation Administration, Aug. 2007.

\bibitem{BibEntry2020Nov(3)}
``14 cfr {\ifmmode\S\else\textsection\fi} 25.203-stall characteristics.'' LII /
  Legal Information Institute, https://www.law.cornell.edu /cfr/text/14/25.203,
  Federal Aviation Administration, Jun. 1995.

\bibitem{BibEntry2020Nov(4)}
``14 cfr {\ifmmode\S\else\textsection\fi} 25.207 - stall warning.'' LII / Legal
  Information Institute, https://www.law.cornell.edu/cfr/text/14/25.207,
  Federal Aviation Administration, Nov. 2014.

\bibitem{agha2017system}
M.~Agha, K.~Kanistras, P.~C. Saka, K.~Valavanis, and M.~Rutherford, ``System
  identification of circulation control uav using x-plane flight simulation
  software and flight data,'' in \emph{AIAA Modeling and Simulation
  Technologies Conference}, Denver, Colorado, Jun. 2017, p. 3154.

\bibitem{bittar2013a}
A.~Bittar and N.~M. de~Oliveira, ``Central processing unit for an autopilot:
  Description and hardware-in-the-loop simulation,'' \emph{Journal of
  Intelligent \& Robotic Systems}, vol.~70, no. 1-4, pp. 557--574, 2013.

\bibitem{meyer2016a}
\emph{X-Plane 11 Desktop Manual | X-Plane}, Laminar Research, 2016.

\bibitem{etin-a}
E.~{\c{C}}etin and A.~T. Kutay, ``Automatic landing flare control design by
  model-following control and flight test on x-plane flight simulator,'' in
  \emph{2016 7th International Conference on Mechanical and Aerospace
  Engineering (ICMAE)}, 2017, pp. 416--420.

\bibitem{junior-a}
J.~M.~M. Junior, T.~Khamvilai, L.~Sutter, and E.~Feron, ``Test platform for
  autopilot system embedded in a model of multi-core architecture using x-plane
  flight simulator,'' in \emph{2019 IEEE/AIAA 38th Digital Avionics Systems
  Conference (DASC)}, 2019, pp. 1--6.

\bibitem{bittar-a}
A.~Bittar, H.~V. Figuereido, P.~A. Guimaraes, and A.~C. Mendes, ``Guidance
  software-in-the-loop simulation using x-plane and simulink for uavs,'' in
  \emph{2014 International Conference on Unmanned Aircraft Systems (ICUAS)},
  2014, pp. 993--1002.

\bibitem{akyurek-a}
S.~Akyurek, G.~Ozden, B.~Kurkcu, U.~Kaynak, and C.~Kasnakoglu, ``Design of a
  flight stabilizer for fixed-wing aircrafts using $h{\infty}$ loop shaping
  method,'' in \emph{2015 9th International Conference on Electrical and
  Electronics Engineering (ELECO)}, 2015, pp. 790--795.

\bibitem{bittar2014a}
A.~Bittar, N.~M.~F. De~Oliveira, and H.~V. De~Figueiredo,
  ``Hardware-in-the-loop simulation with x-plane of attitude control of a suav
  exploring atmospheric conditions,'' \emph{Journal of Intelligent \& Robotic
  Systems}, vol.~73, no. 1-4, pp. 271--287, 2014.

\bibitem{ackerman2014a}
K.~Ackerman, S.~Pelech, R.~Carbonari, N.~Hovakimyan, A.~Kirlik, and I.~M.
  Gregory, ``Pilot-in-the-loop flight simulator for nasa’s transport class
  model,'' in \emph{AIAA Guidance, Navigation, and Control Conference}, 2014,
  p. 0613.

\bibitem{nanduri-b}
A.~Nanduri and L.~Sherry, ``Generating flight operations quality assurance
  (foqa) data from the x-plane simulation,'' in \emph{2016 Integrated
  Communications Navigation and Surveillance (ICNS)}, 2016, pp. 5C1--1.

\bibitem{xplane-b}
\BIBentryALTinterwordspacing
How x-plane works {$\vert$} x-plane. Laminar Research. South Carolina, United
  States. [Online]. Available:
  \url{https://www.x-plane.com/desktop/how-x-plane-works}
\BIBentrySTDinterwordspacing

\bibitem{thong2011a}
C.~W. Thong, ``Modeling aircraft performance and stability on x-plane,''
  \emph{UNSW Canberra ADFA J. Undergrad. Eng. Res}, vol.~3, no.~2, 2010.

\bibitem{pham2014a}
V.~Pham, T.~Bluche, C.~Kermorvant, and J.~Louradour, ``Dropout improves
  recurrent neural networks for handwriting recognition,'' in \emph{2014 14th
  international conference on frontiers in handwriting recognition}, 2014, pp.
  285--290.

\bibitem{zaremba2014a}
W.~Zaremba, I.~Sutskever, and O.~Vinyals, ``Recurrent neural network
  regularization,'' \emph{arXiv preprint arXiv:1409.2329}, 2014.

\bibitem{brochu2010a}
E.~Brochu, V.~M. Cora, and N.~De~Freitas, ``A tutorial on bayesian optimization
  of expensive cost functions, with application to active user modeling and
  hierarchical reinforcement learning,'' \emph{arXiv preprint arXiv:1012.2599},
  2010.

\bibitem{rasmussen2003a}
C.~E. Rasmussen, ``Gaussian processes in machine learning,'' in \emph{Summer
  School on Machine Learning}.\hskip 1em plus 0.5em minus 0.4em\relax Springer,
  2003, pp. 63--71.

\bibitem{kingma2014a}
D.~P. Kingma and J.~Ba, ``Adam: A method for stochastic optimization,''
  \emph{arXiv preprint arXiv:1412.6980}, 2014.

\end{thebibliography}

\end{document}